\documentclass[conference]{IEEEtran}
\IEEEoverridecommandlockouts
\usepackage{cite}
\usepackage{amsmath,amssymb,amsfonts}
\usepackage{algorithmic}
\usepackage{graphicx}
\usepackage{textcomp}
\usepackage{amssymb}
\usepackage{latexsym}
\usepackage{siunitx}

\usepackage{url}
\usepackage{xcolor,colortbl}
\definecolor{newcolor}{rgb}{.8,.349,.1}
\usepackage{gensymb}
\usepackage{todonotes}
\usepackage{xspace} 
\usepackage{multirow}
\usepackage{natbib}
\usepackage{amsmath}
\usepackage{mathtools}
\usepackage{booktabs}
\newcommand{\ours}{UNMuTe\xspace}
\definecolor{light_gray}{gray}{0.95}

\def \ie {\emph{i.e.}}

\usepackage{xcolor}
\def\BibTeX{{\rm B\kern-.05em{\sc i\kern-.025em b}\kern-.08em
    T\kern-.1667em\lower.7ex\hbox{E}\kern-.125emX}}
\begin{document}

\title{UNMuTe: Unifying Navigation and Multimodal Dialogue-like Text Generation\\
}

\author{
    \IEEEauthorblockN{Niyati Rawal, Roberto Bigazzi, Lorenzo Baraldi, Rita Cucchiara}
    \IEEEauthorblockA{University of Modena and Reggio Emilia
    \\\{firstname.surname\}@unimore.it}
}



\maketitle

\begin{abstract}
Smart autonomous agents are becoming increasingly important in various real-life applications, including robotics and autonomous vehicles. One crucial skill that these agents must possess is the ability to interact with their surrounding entities, such as other agents or humans. 
In this work, we aim at building an intelligent agent that can efficiently navigate in an environment while being able to interact with an oracle (or human) in natural language and ask for directions when it is unsure about its navigation performance.
The interaction is started by the agent that produces a question, which is then answered by the oracle on the basis of the shortest trajectory to the goal. The process can be performed multiple times during navigation, thus enabling the agent to hold a dialogue with the oracle. 
To this end, we propose a novel computational model, named UNMuTe, that consists of two main components: a \textit{dialogue model} and a \textit{navigator}. Specifically, the dialogue model is based on a GPT-2 decoder that handles multimodal data consisting of both text and images. First, the dialogue model is trained to generate question-answer pairs: the question is generated using the current image, while the answer is produced leveraging future images on the path toward the goal. Subsequently, a VLN model is trained to follow the dialogue predicting navigation actions or triggering the dialogue model if it needs help.
In our experimental analysis, we show that UNMuTe achieves state-of-the-art performance on the main navigation tasks implying dialogue,~\ie~Cooperative Vision and Dialogue Navigation (CVDN) and Navigation from Dialogue History (NDH), proving that our approach is effective in generating useful questions and answers to guide navigation.
\end{abstract}

\section{Introduction}
In recent years, the advances in Vision-and-Language research have contributed substantially towards the development of the smart embodied agents of the future. 
Aiming to pursue this goal, Vision-and-Language Navigation (VLN)~\citep{anderson2018vision} is a task that lies at an intersection of the three domains of Computer Vision, Natural Language Processing (NLP), and Robotics.
VLN consists of an agent following human instructions while perceiving the environment. However, its standard definition forces the agent to follow textual instructions that are received once and only at the beginning of each episode. This formulation restricts the agent's freedom to interact with the surrounding environment during the duration of the navigation. 
A robot performing VLN is given a natural language sentence in the form ``\textit{Take a right, going past the kitchen into the hallway}'', and can only passively exploit the language modality while retrieving 360$^\circ$ panoramic views of its surroundings. 
%
Engaging in dialogue, instead, 
can aid the agent in successfully navigating unknown environments by asking for help when the trajectory to the goal location is unclear.
The capability to ask questions regarding its current location and where it should move next is a step towards building an intelligent, conversational agent that can communicate and interact with a human while performing intelligent navigation.

Vision-and-Dialogue Navigation (VDN)~\citep{thomason2020vision}, which consists of continuous communication and interaction between an agent and an oracle while performing navigation is the most appropriate candidate to achieve this goal.
However, besides the navigation that is derived from VLN, in VDN some additional aspects need to be addressed: (a) selecting when is the appropriate time to ask a question, (b) deciding which question should be asked, and (c) determine how to answer a given query.
In the task of VDN, no instructions are provided at first but only the name of a target object, however, the agent can query and interact with another agent (the oracle) to gather information on how to navigate in an unseen environment. 
This can also be extended to human-in-the-loop machine learning, where the oracle is a human. 
Nevertheless, most of the previous work in this field does not tackle the generation of the dialogue but performs the navigation task directly training the navigation agent with a human-annotated dialogue between a navigator and an oracle describing the path to a target object. 
Our work differs from these approaches as we train our model to equip a navigation agent with the ability to generate dialogue.

We propose a novel method, called \ours, that consists of two main modules: the first performs navigation or chooses whether to engage in dialogue and the second generates navigation-based dialogue. 
The navigation part consists of a VLN method~\citep{chen2022think} that has been adapted to receive dialogue as input and has been equipped with a policy to decide when to generate dialogues. The dialogue part instead, consists of a Generative Pre-trained Transformer (GPT-2)~\citep{radford2019language} model that is modified to generate pairs of questions and answers conditioned on the target object and the current position of the agent. The connection between these two components is given by a decision mechanism that regulates the generation of dialogue and must be based on the confidence of the navigator. When the navigator is unsure of which direction it has to take, it should ask the oracle for help. 
We compare different dialogue activation policies studying the effect of dialogue generation on navigation.
In our experimental analysis, we prove the effectiveness of the proposed model using the main datasets on VDN~\citep{thomason2020vision}, Cooperative Vision and Dialogue Navigation (CVDN) and Navigation from Dialog History (NDH), and proving that our approach achieves state-of-the-art navigation results on this task. 

\begin{figure}[!t]
\centering
\includegraphics[width=\linewidth]{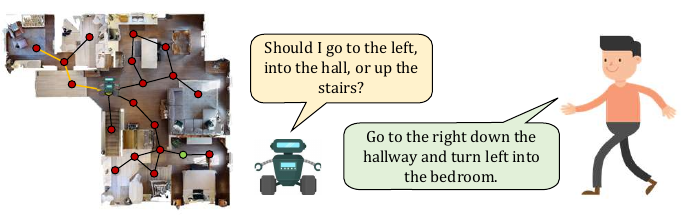}
\caption{We propose a novel computational model that learns to exchange dialogue during navigation when the agent is unsure of the action it should take in the environment. Our proposed model allows the agent to (a) decide when to ask a question, (b) ask target-driven questions, (c) answer given questions, and more importantly, (d) navigate toward the goal.}
\label{fig:overview}
\end{figure}

To summarize, the main contributions of our paper are as follows:
\begin{itemize}
    \item[-] We propose a two-component novel computational model that can both perform navigation following textual instructions in the form of a dialogue and produce question-and-answer pairs that help the navigator to move toward the target object. 
    \item[-] We design a new triggering method involving a learnable threshold used to invoke the generation of question-and-answer pairs when the navigation becomes uncertain.
    \item[-] We perform an extensive experimental analysis to validate the quality of our approach on Cooperative Vision and Dialogue Navigation (CVDN) and Navigation from the Dialogue History (NDH) datasets, showing that our method achieves state-of-the-art performance on goal progress and success rate.
\end{itemize}
\section{Related Work}
\subsection{Vision-and-Language Navigation}
In recent years, research aimed at the development of intelligent autonomous agents has acquired increasing interest with the release of simulation platforms like Gibson~\citep{xia2018gibson}, Matterport3D~\citep{chang2017matterport3d}, and Habitat~\citep{savva2019habitat}, as well as datasets enabling object interaction~\citep{shridhar2020alfred,padmakumar2022teach,gao2022dialfred}. Among the various embodied tasks that are the object of this research line, Vision-and-Language Navigation (VLN) aims to implement such agents with multimodal reasoning capabilities in both indoor and outdoor environments. In fact, VLN requires an agent to interpret human instructions, in the form of natural language text, while perceiving observations of the environment. Among indoor VLN methods,
~\cite{anderson2018vision} first tackled the task by adopting sequence-to-sequence long short-term memories for action inference.
~\cite{fried2018speaker} started exploiting the panoramic observation space and introduced a module for synthetic instructions generation.
~\cite{fu2020counterfactual} instead, used counterfactual thinking to perform data augmentation. More recently,
~\cite{ma2019self, ma2019regretful} proposed a model with a self-monitoring agent, and
~\cite{landi2019embodied} used dynamic convolution filters. RCM~\citep{wang2019reinforced} employed a reinforcement learning training approach to improve cross-modal matching and
~\cite{hong2020language} implemented graphs to model relations between scenes, objects, and instructions. More recently, Transformer-based~\citep{vaswani2017attention} models have become popular. Among these approaches, VLN$\circlearrowright$ BERT~\citep{hong2021vln} implemented a recurrent BERT~\citep{devlin2018bert} to model time dependencies, while PTA~\citep{landi2021multimodal} and HAMT~\citep{chen2021history} used Transformers to respectively perform multimodal fusion and exploit episode history. Topological maps and a dual-scale Transformer are proposed by
~\cite{chen2022think} to consider both long-term action planning and fine-grained understanding. In our approach, the navigation module uses a modified version of DUET to select the nodes visited by the agent. In contrast to the setting we tackle in this work, VLN does not allow the exchange of textual information besides the human instructions at the beginning of each episode. Some methods for VLN that tried to address this lack, are proposed by
~\cite{nguyen2019help} and
~\cite{chi2020just}. However,
~\cite{nguyen2019help} used preset language-assisted routes, and 
~\cite{chi2020just} limited the agent interaction to only one possible question and the response given by the oracle is the next action on the shortest path route to the goal, whereas our approach only exchanges textual information. Moving on to outdoor VLN approaches, the agent has to perform navigation in an urban environment where the visual appearance is more repetitive and clear landmarks are difficult to be found. While StreetLearn~\citep{mirowski2018learning} is the first dataset providing panoramic views of the streets of Manhattan and Pittsburg for navigation, it does not provide human-annotated instructions but only provides directions and street names toward the target location. Touchdown dataset~\citep{chen2019touchdown,mehta2020retouchdown} introduces human instructions for a subset of the StreetLearn dataset. Another large-scale dialogue dataset is called “Talk The Walk”~\citep{de2018talk} and involves two agents (a “guide” and a “tourist”) that communicate in natural language to achieve a common goal.

%

\subsection{Vision-and-Dialogue Navigation}
Constraining the navigation in VLN to follow human instructions that are given only at the beginning of each episode could lead the agent to diverge from the correct trajectory when the match between instruction and visual cues is not clear. 
In this context, extending the task by allowing the agent to generate conversations with an oracle asking for new instructions could redirect the agent in the correct direction to the goal.
However, this relaxation of the VLN task introduces new challenges defined by the generation of an appropriate question and by the decision of the most suitable moment for such interaction.
The benchmark used to evaluate dialogue-based agents is defined by the contribution of
~\cite{thomason2020vision}, which introduced Cooperative Vision and Dialogue Navigation (CVDN), a dataset of over $2$K embodied trajectories with human-human dialogues in the simulated indoor environments of Matterport3D~\cite{Matterport3D}, and Navigation from Dialog History (NDH), a task of $7$K navigation episodes using CVDN dialogues as textual input. 
In particular, the CVDN dataset is annotated using two humans, a navigator and an oracle, where the first has to navigate toward a predefined target object while being able to ask the oracle for directions, and the oracle can access the shortest path trajectory from the current position of the navigator to the target.
However, most of the existing studies tackling VDN use the dialogue only as an input for the navigation method \citep{anderson2018vision,hao2020towards,zhu2020vision,chen2021history,qiao2022hop,zheng2023esceme}. In these approaches, the agent does not generate dialogue. 
On the contrary, RMM \citep{roman2020rmm} designed three agents, two of them are entitled of producing a dialogue aimed at a target object regularly, while the third is in charge of the navigation.
~\cite{zhu2021self} proposed a computational model that engages in dialogue only when the navigating agent is unsure of which action to take. However, the generated dialogue is based on textual templates and consists of questions that have affirmative or negative answers, with the navigation agent that is rewarded for producing questions that have ``yes'' as the answer. Yet another work introduces a model VISITRON that learns when to navigate and when to ask questions \cite{shrivastava2021visitron}. In contrast to these methods, we propose a purely generative speaker model that produces elaborated conversations with detailed answers. Additionally, the agent also has to decide when to engage in dialogue.

\subsection{Text Generation for Visual Navigation}
The idea of generating synthetic text for visual navigation has arisen naturally from the goal of improving the performance of a VLN agent. In fact, from the early work on VLN, a specific line of research focused on augmenting human-annotated datasets with well-formed synthetic instructions~\citep{fried2018speaker,majumdar2020improving}. For example,
~\cite{zhu2020multimodal} converted the instructions provided by the Google Maps API in the StreetLearn dataset to human-like instructions using a text-style transfer approach, showing improvements for outdoor VLN agents. Another line of research uses speaker models to generate textual instructions using sequences of images belonging to navigation trajectories. This framework can also be extended to unlabelled environments, as shown by
~\cite{chen2022learning}. Synthetically augmented datasets have been proven to improve the performance of navigation agents on several VLN datasets~\citep{fried2018speaker,majumdar2020improving,guhur2021airbert,wang2021less,chen2022learning,zhu2020multimodal}. An evolution of this idea would be equipping navigation agents with the ability to produce conversations aimed at the target location or object. 

In our approach, we exploit a speaker model that generates question-and-answer pairs conditioned on the trajectory to CVDN and NDH targets, and we use the generated dialogue to guide a navigation agent.

\section{Proposed Method}
\begin{figure*}[!t]
\centering
\includegraphics[width=.79\linewidth]{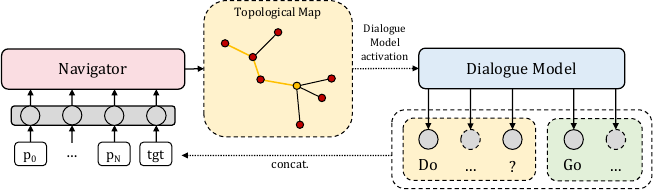}
\caption{\ours consists of a dialogue model that is based on a GPT-2 decoder and a navigation model that is based on a state-of-the-art navigator, \ie~DUET~\citep{chen2022think}. When DUET is unsure of the action the agent should take, it outputs an action that prompts the dialogue model to generate a question and an answer regarding where the agent should move.}
\label{fig:architecture}
\end{figure*}

We propose a novel computational model called \ours (visually depicted in Fig.~\ref{fig:architecture} and~\ref{fig:dialogue_model}), which is composed of a navigation model that predicts the actions of the agent and a dialogue model that, when triggered, generates question-and-answer pairs denoting the trajectory to the goal. First, the dialogue model is trained individually such that the model can generate questions and answers. Next, the navigator model is trained with the help of the dialogue model. Specifically, the navigator model can consult the dialogue model when it is confused regarding which action to take. 
Given the current observation of the navigator and the target object, the dialogue model generates a question and an answer conditioned on the trajectory to the target. The navigator model uses the output of the dialogue model to select its next action thereby improving the final navigation performance. 

\subsection{Dialogue Model}
\begin{figure*}[!t]
\centering
\includegraphics[width=.6\linewidth]{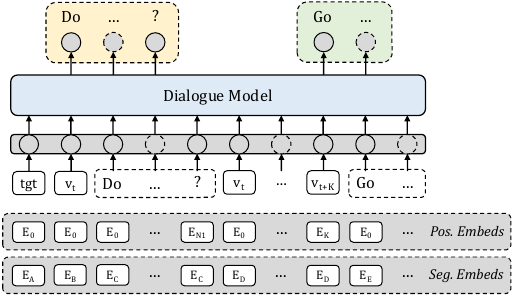}
\caption{Dialogue model with corresponding inputs and outputs. The model is trained to predict the subsequent language token belonging to the sequence. To facilitate graphical presentation special tokens such as \texttt{BOS} or \texttt{EOS} are omitted.}
\label{fig:dialogue_model}
\end{figure*}

The dialogue model, shown in Fig.~\ref{fig:dialogue_model}, is a single Generative Pre-trained Transformer (GPT-2) that generates question-and-answer pairs starting from the target object and the current observation of the agent. Inspired by
~\cite{alayrac2022flamingo}, the dialogue model is finetuned conditioning on visual inputs to achieve multimodal capabilities using the trajectories and the conversations contained in the CVDN dataset. 
The actual input of the dialogue model can be split into three components: the token of the target object label, the image features and textual tokens associated with the question, and the image features and textual tokens associated with the answer. Formally,
\begin{equation}
    \begin{split}
    y = \texttt{GPT}\bigg(\bigg[
    \texttt{BOS},
    \underbracket{o_\text{tgt}}_{\text{Target}},
    \texttt{EOS},
    v_t,
    \texttt{BOS},
    \underbracket{
        q_1,.., q_n,
    }_{\text{Question}}
    \texttt{EOS}, \\
    v_{t},.., v_{t+k},
    \texttt{BOS},
    \underbracket{ 
        a_1,.., a_m,
    }_{\text{Answer}}
    \texttt{EOS}
    \bigg]\bigg)
    \label{eq:gpt}
    \end{split}
\end{equation}
where $o_\text{tgt}$ indicates the target object label, $\texttt{SEP}$ is a separator token, $v_t$ the visual features related to the current observation of the agent, $( q_1, ..., q_n )$ the actual question tokens. Correspondingly, $( v_{t}, ..., v_{t+k} )$ denotes the set of visual features and $( a_1, ..., a_m )$ the tokens corresponding to the answer.

All the image features used for the dialogue model are extracted using a pretrained visual encoder. During training, the dialogue model learns to predict the subsequent language token of both the question and the answer, starting from the \texttt{BOS} token. Instead, all the tokens following image features are ignored. The generation of the question is influenced only by the current observation of the agent, while the answer is conditioned with $k$ additional observations that are collected along the trajectory to the target.
The trajectory to the goal is obtained using Dijkstra's algorithm on the navigation graph between the current node and the target node.

In addition to the token embeddings, the proposed dialogue model uses position and segment embeddings to effectively segregate the information regarding the different components and modalities of the input.
This choice was inspired by
~\cite{devlin2018bert}. 
During inference, the output of the dialogue model is generated token by token autoregressively until the \texttt{EOS} token of the generated answer is produced.

\subsection{Navigator Model}
The navigator model consists of a modified variant of Dual Scale Graph Transformer (DUET)~\citep{chen2022think}. 
DUET keeps track of visited and observed nodes by producing a topological map of the environment. At each time step, the map is updated storing the visual features associated with newly visited nodes and navigable nodes.
Graph Transformers are used to combine a fine-scale encoding over the local observations and a coarse-scale encoding on the global map.

However, the original architecture of DUET prohibits backtracking by masking out visited nodes in the action space. While this implementation holds when following the shortest trajectory from a certain position to the goal, it fails when the supervision is performed using human-generated trajectories as in CVDN, as they could contain backtracks. Therefore, revisiting the same node multiple times might be necessary. We modify DUET accordingly to account for this behavior. Originally, DUET masks all the nodes previously visited to prevent the agent from revisiting these nodes. We remove the masking of previously visited nodes and only mask the current node so to ensure that the agent does not remain on the same node.

The prediction of the next location, after this modification, considers an action space comprising all the possible navigable nodes in the graph instead of only the neighboring ones. Additionally, the action space includes an additional possibility defined by the stop action. 
As in CVDN the only available textual input at the beginning of the episode is the target object, we mimic an instruction including such object by prepending learnable prompt embeddings at the beginning of the input to the model. 

\subsection{Dialogue Exchange during Navigation}
\label{sec:entr_thresh}
As represented in Fig.~\ref{fig:architecture}, \ours comprises of a dialogue model and a navigation model, where the navigation model can trigger the dialogue model to generate a question-and-answer pair when the trajectory to the target is not clear. 
In this respect, the confidence of the navigator can be quantified as the entropy $\mathcal{H}$ of its action probability distribution, which acquires higher values as the probability distribution approaches the uniform distribution.
%
Therefore, the entropy $\mathcal{H}$ of the action probability distribution over the navigable nodes of the environment is computed at each time step. When the entropy $\mathcal{H}_t$ exceeds a threshold value $\alpha$, the navigator triggers the dialogue model and the dialogue generation is activated. The conversation returned by the dialogue pair is concatenated to the input of the navigator to recompute the probability distribution over the action space, and if $\mathcal{H}_{t+1}\leq \alpha$, the next viewpoint is selected for the navigation.

We perform an empirical analysis of the choice of the entropy threshold and evaluate the use of a learnable parameter $\hat{\alpha}$ as threshold. 
To this end, a binary cross-entropy loss is used to set a threshold value $\hat{\alpha}$ which is higher than the entropy in the nodes of the graph where the dataset contains dialogue annotations, and is lower otherwise:
\begin{equation}
    \mathcal{L}_{QA} = \textnormal{BCE}(q, \bar{q}), \quad \textnormal{s.t.} \quad q=\frac{1}{1+e^{(\hat{\alpha}-\mathcal{H}_t)}}
    \label{eq:qa_loss}
\end{equation}
where $\bar{q}$ is 1 if a question is asked at time step $t$, and 0 otherwise. 
As the training of DUET is done using both teacher forcing,~\ie~following the ground truth trajectory, and by sampling from the action probability distribution, we calculate $\mathcal{L}_{QA}$ only for the teacher forcing training stage.
When the actions are sampled, the value of $q$ in Eq.~\ref{eq:qa_loss} is only used to trigger the dialogue model.

\begin{figure*}[!t]
    \centering
    \resizebox{.7\linewidth}{!}{
        \begin{tabular}{cc}
            \includegraphics[height=0.48\textwidth]{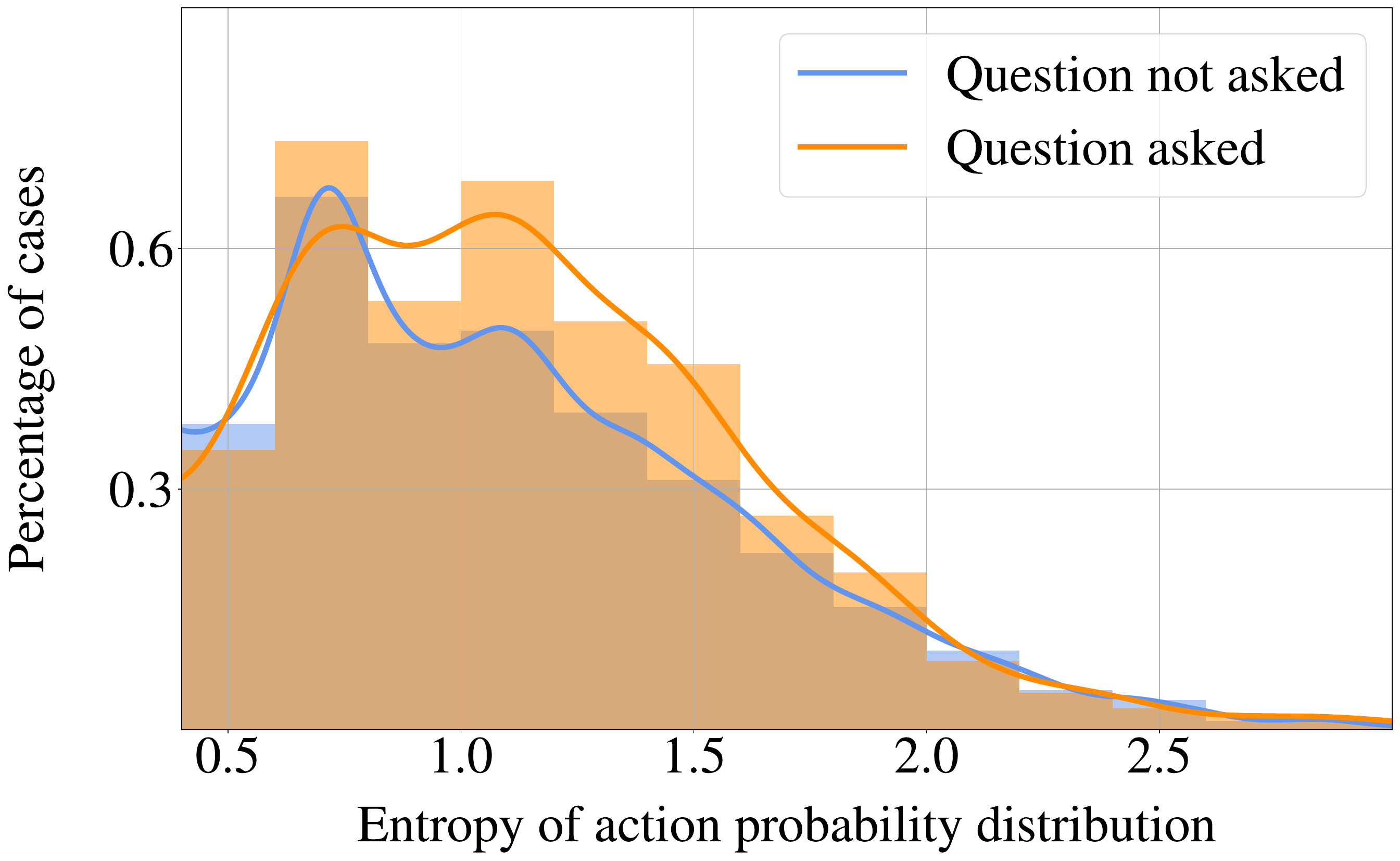} &
            \includegraphics[height=0.48\textwidth]{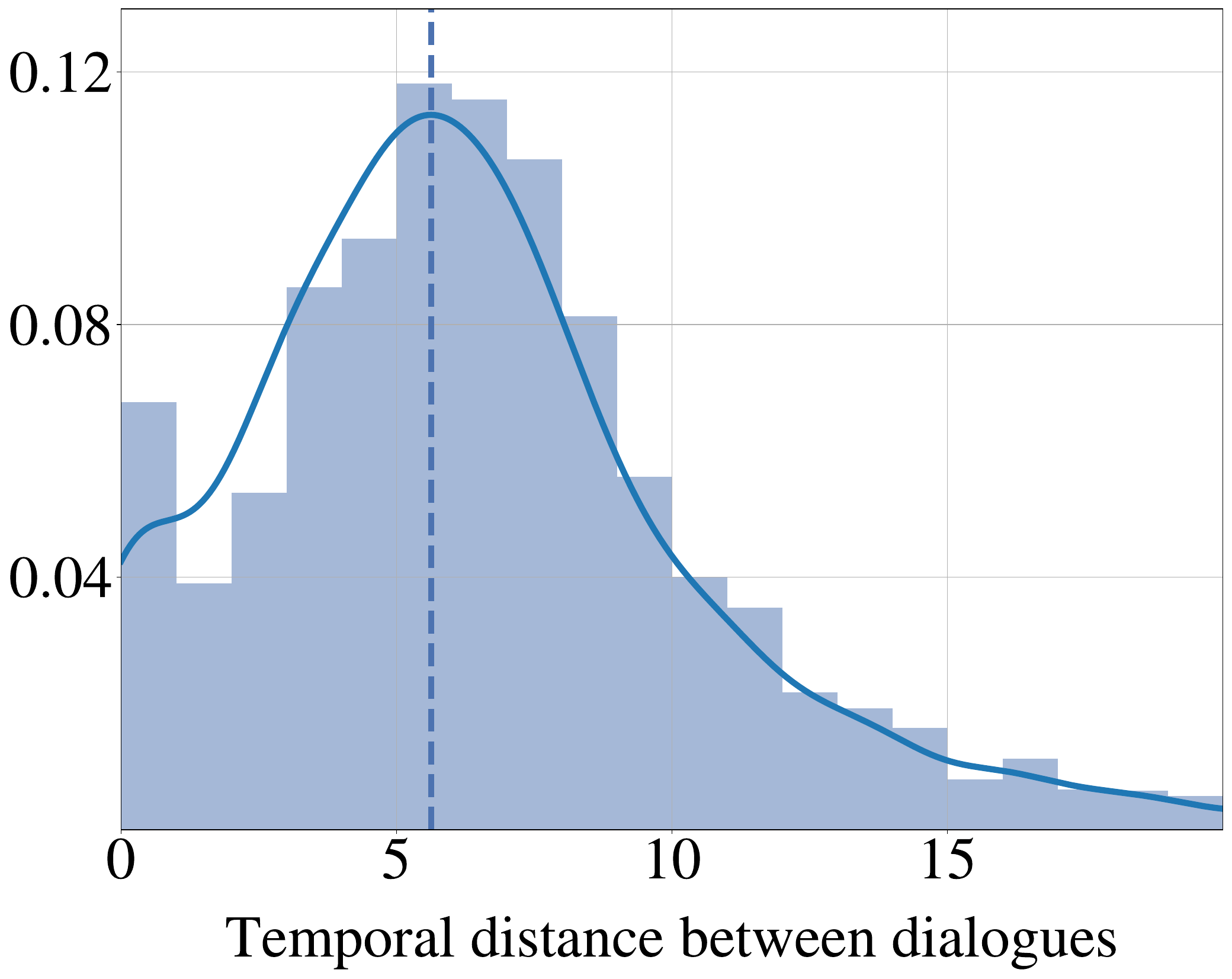} \\
        \end{tabular}
    }
    \caption{Probability distributions of the entropy of the action probability and the temporal distances between dialogues on the training split of CVDN.}
    \label{fig:plots}
\end{figure*}

\begin{table}[!t]
\centering
\caption{Hyperparameters related to the navigator model.}
\resizebox{.7\linewidth}{!}{
\begin{tabular}{p{6cm} cc}
\toprule
\multicolumn{2}{c}{\textbf{Navigator}} \\ 
\midrule
num text encoder layers: & $9$ \\
\rowcolor{light_gray} 
num coarse-scale encoder layers: & $4$ \\
num fine-scale encoder layers: & $4$ \\
\rowcolor{light_gray} 
num pano layers:  & $2$ \\
max action length: & $15$ \\
\rowcolor{light_gray} 
max instruction length: & $512$  \\
training batch size: & $2$  \\
\rowcolor{light_gray} 
learning rate: & $10^{-5}$ \\
sample weight: & $1.0$  \\
\rowcolor{light_gray} 
ml weight: & $0.2$  \\
\bottomrule
\end{tabular}
}
\label{tab:hyperparams_nav}
\end{table}

\section{Experiments}

\begin{table}[!t]
\centering
\caption{Hyperparameters related to the dialogue model.}
\resizebox{.7\linewidth}{!}{
\begin{tabular}{p{6cm}c  c}
\toprule
\multicolumn{2}{c}{\textbf{Dialogue Model}} \\
\midrule
num layers: & $12$\\
\rowcolor{light_gray} 
model dimensionality: & $768$\\
num attention heads: & $12$\\
\rowcolor{light_gray} 
training batch size: & $12$\\
learning rate: & $10^{-4}$\\
\rowcolor{light_gray} 
max instruction length: & $1024$ \\
num imgs used to generate question: & $1$ \\
\rowcolor{light_gray} 
num future imgs to generate answer: & $20$\\
optimizer: & adam \\
\bottomrule
\end{tabular}
}
\label{tab:hyperparams_dialogue}
\vspace{-4mm}
\end{table}

\begin{table*}[t]
\centering
\caption{Navigation results for our approach and recent methods on the ``val unseen'' split of CVDN. 
} 
\resizebox{.6\linewidth}{!}{
\begin{tabular}{p{8cm} | cccc }
\toprule
\multicolumn{1}{c}{} & \multicolumn{4}{c}{\textbf{Val Unseen}} \\
\addlinespace[4pt]
& GP & SPL & SR & nDTW \\
\midrule
RMM$_{n=3}$ + Oracle Stopping~\citep{roman2020rmm} & 8.9 & - & - & -  \\
\rowcolor{light_gray} 
SCoA~\citep{zhu2021self} & 11.19 & - & - & -  \\
\midrule
\textbf{\ours (threshold)} & \textbf{13.35} & 5.39 & 7.31 & 24.81  \\
\rowcolor{light_gray} 
\textbf{\ours (4 time steps)} & 12.68 & 3.62 & 5.00 & 24.44 \\
\textbf{\ours (5 time steps)} & 13.13 & \textbf{7.73} & \textbf{9.62} & \textbf{25.76}  \\
\rowcolor{light_gray} 
\textbf{\ours (6 time steps)} & 12.31 & 4.81 & 5.77 & 23.65  \\
\bottomrule
\end{tabular}
} 

\label{tab:cvdn_val_unseen}
\vspace{-1mm}
\end{table*}

\begin{table}[!t]
\centering
\caption{Comparison of navigation results with different image feature extractors on CVDN val unseen.}
\resizebox{.88\linewidth}{!}{
\begin{tabular}{p{4.5cm} | cccc }
\toprule
\multicolumn{1}{c}{} & \multicolumn{4}{c}{\textbf{Val Unseen}} \\
\addlinespace[4pt]
& GP & SPL & SR & nDTW \\
\midrule
\textbf{\ours (BLIP)} & 12.05 & 4.97 & 6.54 & 21.67 \\
\rowcolor{light_gray} 
\textbf{\ours (ViT-L/16)} & 12.29 & 5.99 & \textbf{8.46} & 24.78 \\
\textbf{\ours (CLIP ViT-L/14)} & 12.21 & {4.78} & {6.15} & {23.78} \\
\rowcolor{light_gray} 
\textbf{\ours (CLIP RN50)} & 11.83 & 4.94 & 6.15 & \textbf{25.11} \\
\textbf{\ours (ResNet50)} & 12.34 & \textbf{6.68} & 8.08 & 23.80  \\
\rowcolor{light_gray} 
\textbf{\ours (ResNet152)} & \textbf{13.35} & 5.39 & 7.31 & 24.81 \\
\bottomrule
\end{tabular}
} 

\label{tab:feat}
\vspace{-1mm}
\end{table}

\begin{table}[!t]
\centering
\caption{Comparison of navigation results with different constant thresholds on CVDN val unseen. 
}
\resizebox{.88\linewidth}{!}{
\begin{tabular}{p{4.5cm} | cccc }
\toprule
\multicolumn{1}{c}{} & \multicolumn{4}{c}{\textbf{Val Unseen}} \\
\addlinespace[4pt]
& GP & SPL & SR & nDTW \\
\midrule
\textbf{\ours (w/o prompts)} & 11.97 & \textbf{8.12} & \textbf{10.77} & \textbf{25.48}  \\
\rowcolor{light_gray} 
\textbf{\ours (4 prompts)} & \textbf{13.35} & 5.39 & 7.31 & 24.81  \\
\textbf{\ours (8 prompts)} & 11.96 & 8.49 & 11.92 & 23.30  \\
\bottomrule
\end{tabular}
} 
\label{tab:prompts}
\vspace{-2mm}
\end{table}

\begin{table}[t]
\centering
\caption{Comparison of navigation results with different numbers of prompt embeddings on CVDN val unseen. 
}
\resizebox{.88\linewidth}{!}{
\begin{tabular}{p{4.5cm} | cccc }
\toprule
\multicolumn{1}{c}{} & \multicolumn{4}{c}{\textbf{Val Unseen}} \\
\addlinespace[4pt]
& GP & SPL & SR & nDTW \\
\midrule
\textbf{\ours (learnable thr.)} & \textbf{13.35} & 5.39 & 7.31 & \textbf{24.81} \\
\rowcolor{light_gray} 
\textbf{\ours (thresh=0.9)} & 12.99 & \textbf{6.99} & \textbf{9.62} & 24.21  \\
\textbf{\ours (thresh=1.0)} & 11.27 & 5.28 & 6.92 & 22.14  \\
\rowcolor{light_gray}
\textbf{\ours (thresh=1.1)} & 12.03 & 5.62 & 7.31 & 23.45  \\
\bottomrule
\end{tabular}
} 

\label{tab:fixed}
\vspace{-2mm}
\end{table}

\subsection{Experimental Setup}
We evaluate the effectiveness of UNMuTe on Vision-and-Dialog Navigation (VDN) using both CVDN and NDH datasets. CVDN contains 2050 navigation trajectories performed on a total of 83 environments of Matterport3D~\citep{chang2017matterport3d}, while NDH is composed of 7K navigation episodes obtained by splitting CVDN trajectories in multiple instances. The navigation episodes are performed on navigation graphs where each node is defined by a 360° RGB observation. Even if the navigation module exploits the complete panoramic image to compute its output, the dialogue model uses only frontal crops of 60° to generate the conversation pairs forcing the generated text to refer to the scene in the direction of the agent. In Tab.~\ref{tab:hyperparams_nav} and Tab.~\ref{tab:hyperparams_dialogue}, we show the most relevant hyperparameter values used to implement the models composing \ours.
For the GPT-2 decoder, we use a medium-sized, pre-trained version with L = 12, d = 768, H = 12, where L is the number of layers, d is the model dimensionality, and H is the number of attention heads. The resulting dialogue model contains 124M parameters and was trained for approximately 6 hours. The navigation model (164M parameters) was finetuned for 48 hours each on a single NVIDIA RTX6000 GPU. The visual features used by UNMuTe are extracted using ResNet-152 model. 
The experimental results contained in this section are compared with the current state-of-the-art methods on both CVDN and NDH datasets. While the evaluation using NDH dataset is more popular, interactive experiments on CVDN are only performed by RMM~\citep{roman2020rmm} and SCoA~\citep{zhu2021self}. RMM uses two speaker models that regularly generate questions and answers, while SCoA uses a model to predict when to generate dialogue and selects the most appropriate question among a set of question templates. The main competitor on NDH are instead, HAMT~\citep{chen2021history} and VISITRON~\citep{shrivastava2021visitron}. HAMT encodes episode history and uses it as an additional modality with text and images to predict its actions, while VISITRON trains a multimodal Transformer encoder and an LSTM decoder to predict navigation actions and when to exchange dialogue. 

The metrics employed for the navigation experiments are goal progress (GP), \ie~the mean reduction in Euclidean distance between the starting position and to final position with respect to the target; success rate (SR), \ie~the fraction of episodes where the agent can reach the goal position within $3$ meters; success rate weighted by path length (SPL); and normalized Dynamic Time Warping (nDTW) as defined by
~\cite{ilharco2019general}.

\subsection{CVDN Experiments} 
The experiments performed on the CVDN dataset are presented in Tab.~\ref{tab:cvdn_val_unseen} and showcase the quality of the overall approach in an interactive setting. In fact, during the navigation using the episodes of CVDN, the model has to autonomously trigger the dialogue model to generate question-and-answer pairs to guide its movement toward the target.

We compare different configurations of \ours, using the learnable threshold presented in Sec.~\ref{sec:entr_thresh}, and a policy that activates at regular time intervals. The latter is obtained on the basis of the distribution of the training split of CVDN (shown in Fig.~\ref{fig:plots}), by considering the mode of the temporal distance between ground-truth dialogues.
%
%
As the mode of the temporal distances distribution is 5.63, we generate question-and-answer pairs every 4, 5, and 6 time steps during training and evaluation on the CVDN task. 
Triggering the dialogue model every $5$ time steps achieves a state-of-the-art success rate of $7.73$ and SPL of $9.62$. State-of-the-art goal progress of $13.35$ meters is obtained by the model with a learnable entropy threshold, thus confirming the effectiveness of this strategy. We also compare \ours with the current state-of-the-art methods, which, however, do not evaluate in terms of SPL, SR, and nDTW, but only present GP results. 
All configurations of \ours present better results than the competitors, with the best configuration that overcomes SCoA by $2.16$ meters in terms of goal progress.

\begin{table*}[t]
\centering
\caption{Navigation metrics for our approach and competitors on the ``val unseen'' and ``test unseen'' splits of the NDH dataset. 
\label{tab:ndh_val_unseen}
}
\resizebox{.66\linewidth}{!}{
\begin{tabular}{p{8cm} | ccc | ccc}
\toprule
\multicolumn{1}{c}{} & \multicolumn{3}{c}{\textbf{Val Unseen}} & \multicolumn{3}{c}{\textbf{Test Unseen}} \\
\addlinespace[4pt]
& GP & SPL & SR & GP & SPL & SR  \\
\midrule
Seq2Seq \citep{anderson2018vision} & 2.10 & - & - & 2.35 & 16 & -  \\
\rowcolor{light_gray} 
PREVALENT \citep{hao2020towards} & 3.15 & - & - & 2.44 & 24 & -  \\
CMN \citep{zhu2020vision}  & 2.97 & - & -  & 2.95 & 1 & -  \\
\rowcolor{light_gray} 
HOP \citep{qiao2022hop} & 4.41 & - & - & 3.24 & - & -  \\
HAMT \citep{chen2021history} & 5.13 & - & - & 5.58 & 7 & -  \\
\midrule
ScoA \citep{zhu2021self} &  2.91 & - & - & 3.37 & 15 & -  \\
\rowcolor{light_gray} 
VISITRON \citep{shrivastava2021visitron} & 3.25 & 11 & 27 & 3.11 & 12 & - \\
VISITRON (Best SPL) \citep{shrivastava2021visitron} & 2.71 & 25 & 33 & 2.40 & 25 & - \\
\midrule
\textbf{\ours (Planner)} & 4.98 & \textbf{49} & \textbf{60} & 4.03 & \textbf{47}  & \textbf{56} \\
\rowcolor{light_gray} 
\textbf{\ours (Player)} & \textbf{5.88} & 22 & 36 & \textbf{5.75}  & 22 & 35 \\
\bottomrule
\end{tabular}
} 

\vspace{1mm}
\end{table*}

\begin{table*}[t]
\centering
\caption{Evaluation in terms of text generation quality. 
}
\resizebox{.61\linewidth}{!}{
\setlength{\tabcolsep}{0.3em}
\begin{tabular}{p{5.5cm} | ccccc}
\toprule
\multicolumn{1}{c}{} & \multicolumn{5}{c}{\textbf{Val Unseen}} \\
\addlinespace[4pt]
& BLEU-1 & METEOR & ROUGE & CIDEr & SPICE \\
\midrule
{\textbf{Questioner}} & 0.201 & 0.092 & 0.179 & 0.181 & 0.089 \\
\midrule
{\textbf{Oracle w/o future images}} & 0.214 & 0.091 & 0.177 & 0.111 & 0.088 \\
\rowcolor{light_gray} 
{\textbf{Oracle w/o target object}} &  0.228 & 0.098 & 0.192 & 0.145 & 0.094 \\
{\textbf{Oracle}} & 0.237 & 0.098 & 0.200 & 0.179 & 0.109 \\
\bottomrule
\end{tabular}
}
\label{table:description_metrics}
\vspace{-1mm}
\end{table*}

\textbf{Experiments using Different Extracted Image Features.} 
In Tab.~\ref{tab:feat}, we selected the most appropriate pretrained visual encoder for the extraction of the image features for our dialogue model assessing the results of different models: ResNet152~\citep{he2016deep}, ResNet50~\citep{he2016deep}, CLIP ~\citep{radford2021learning}, BLIP~\citep{li2022blip} and ViT-L/16~\citep{dosovitskiy2021image}. In the case of CLIP, we consider the variants exploiting ViT-L/14 and RN50 as backbones. Following previous work on Vision-and-Dialog Navigation, we prioritized models with better goal progress and found out that the navigation results of the agent using image features extracted with ResNet-152 achieved the best performance. 
The goal progress for \ours using ResNet152 features is better than the other configurations by at least $1.01$ meters.

\textbf{Experiments using Different Prompt Embedding Sizes.}
We performed an ablation study on the navigation performance of \ours using different numbers of learnable prompt embeddings at the beginning of the instruction used by the navigator. 
We compared a model not using learnable prompt embeddings with models using respectively $4$ and $8$ learnable prompt embeddings.
For all the navigators considered in this experiment, the questions were asked using the learnable entropy threshold. As we can see in Tab. \ref{tab:prompts}, \ours with 4 learnable prompts has the best performance in terms of goal progress (GP) with an increase of $1.39$ meters over \ours with $8$ prompt embeddings and $1.38$ meters over the model that does not use prompt embeddings.

\begin{figure*}[!t]
    \centering
    \small
    \resizebox{.95\linewidth}{!}{
        \setlength{\tabcolsep}{.06em}
        \begin{tabular}{cccccccc}
            \includegraphics[width=0.1\textwidth]{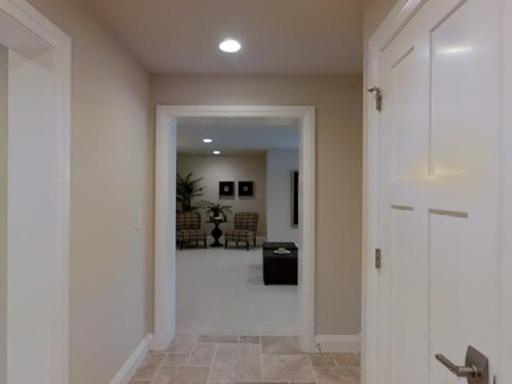} &
            \includegraphics[width=0.1\textwidth]{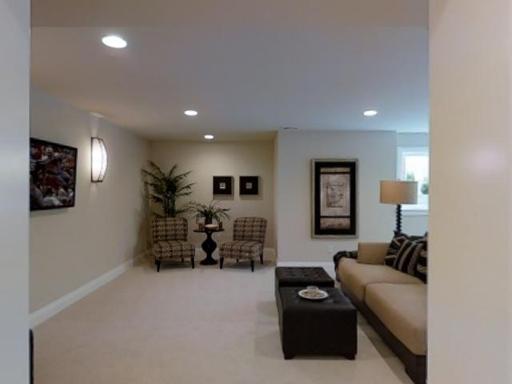} &
            \includegraphics[width=0.1\textwidth]{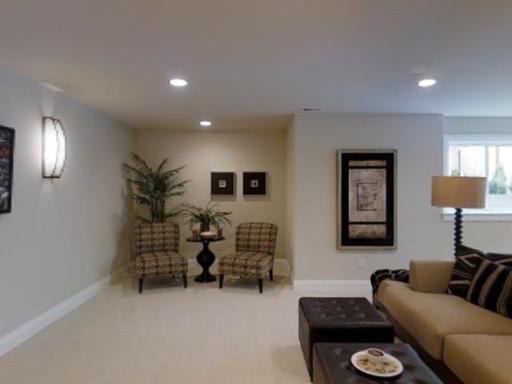} &
            \includegraphics[width=0.1\textwidth]{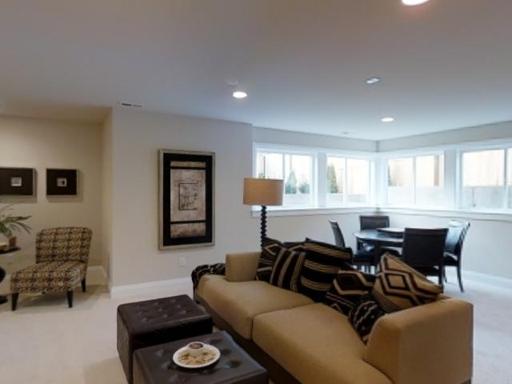} &
            \includegraphics[width=0.1\textwidth]{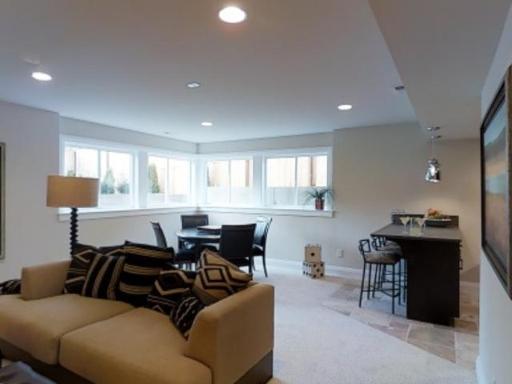} &
            \includegraphics[width=0.1\textwidth]{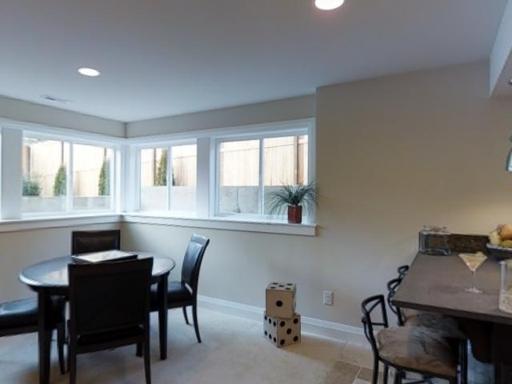} &
        \end{tabular}
    }
    \resizebox{.9\linewidth}{!}{
    \centering
        \begin{tabular}{cp{140mm}}
           \addlinespace[-2pt]
           \multirow{2}{2cm}{\raggedleft \textbf{GT}:} & \textit{Should i go back down this hall?}
           \newline
           \textit{It wants you to make a left turn and go in that family room} \\
           \addlinespace[4pt]
           \multirow{2}{2cm}{\raggedleft \textbf{UNMuTe}:} & \textit{Which way from here?}
           \newline
           \textit{Make a right and go towards the living room.} \\
           \addlinespace[10pt]
        \end{tabular}
    }
    \resizebox{.95\linewidth}{!}{
        \setlength{\tabcolsep}{.06em}
        \begin{tabular}{cccccccc}
            \includegraphics[width=0.1\textwidth]{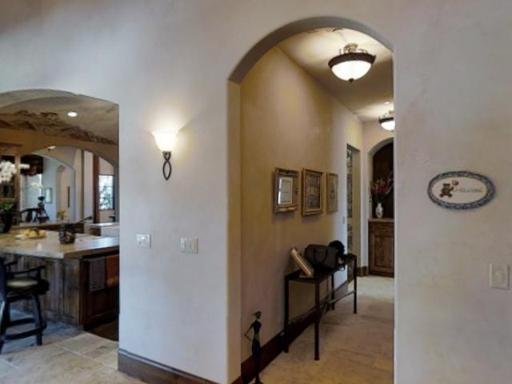} &
            \includegraphics[width=0.1\textwidth]{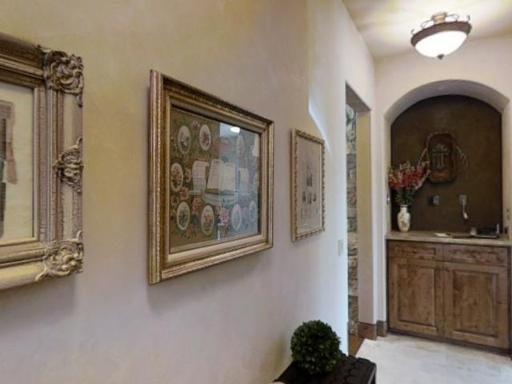} &
            \includegraphics[width=0.1\textwidth]{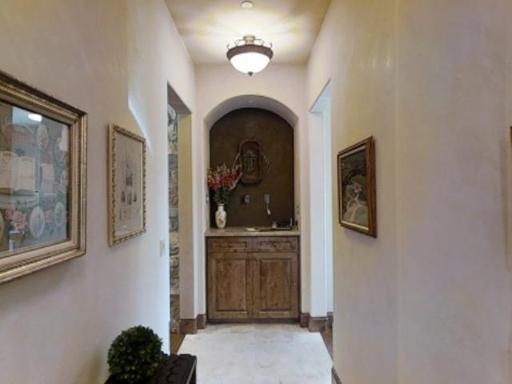} &
            \includegraphics[width=0.1\textwidth]{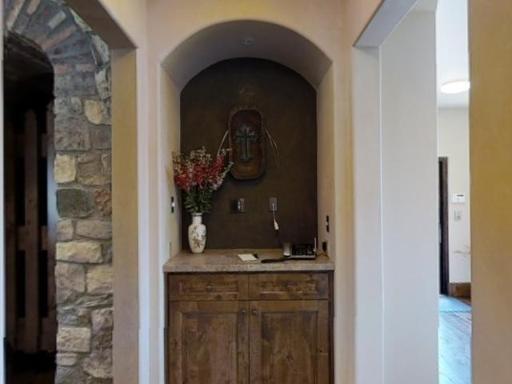} &
            \includegraphics[width=0.1\textwidth]{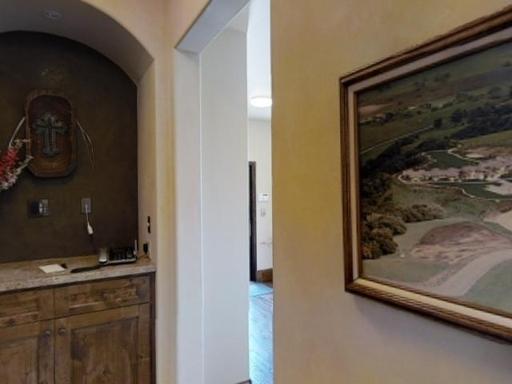} &
            \includegraphics[width=0.1\textwidth]{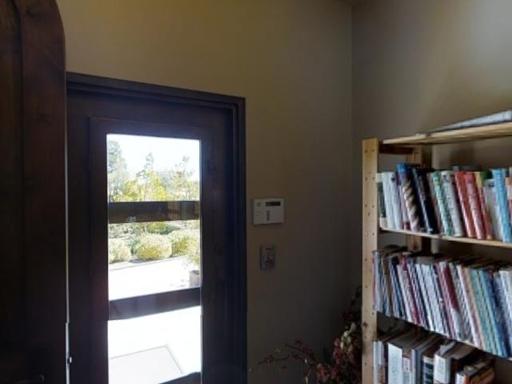} &
        \end{tabular}
    }
    \resizebox{.9\linewidth}{!}{
    \centering
        \begin{tabular}{cp{140mm}}
           \addlinespace[-2pt]
           \multirow{2}{2cm}{\raggedleft \textbf{GT}:} & \textit{Lt straight or rt?}
           \newline
           \textit{Turn right, then all the way down the hallway, there will be a room at the end of the hallway on the right.} \\
           \addlinespace[4pt]
           \multirow{2}{2cm}{\raggedleft \textbf{UNMuTe}:} & \textit{Do I go down the long hallway here? }
           \newline
           \textit{Yes, go down the long hall to the living room} \\
           \addlinespace[10pt]
        \end{tabular}
    }
    \resizebox{.95\linewidth}{!}{
        \setlength{\tabcolsep}{.06em}
        \begin{tabular}{cccccccc}
            \includegraphics[width=0.1\textwidth]{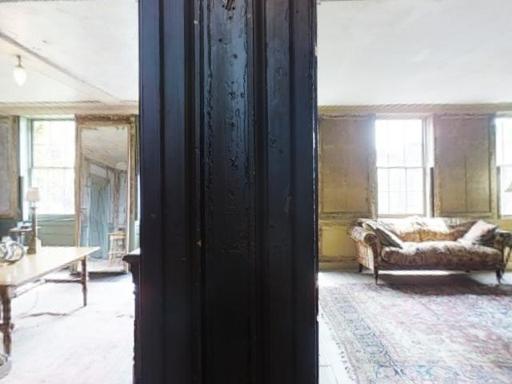} &
            \includegraphics[width=0.1\textwidth]{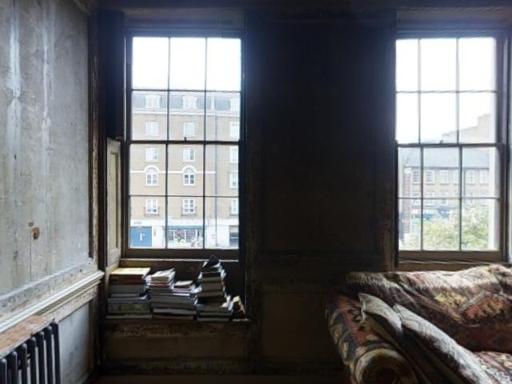} &
            \includegraphics[width=0.1\textwidth]{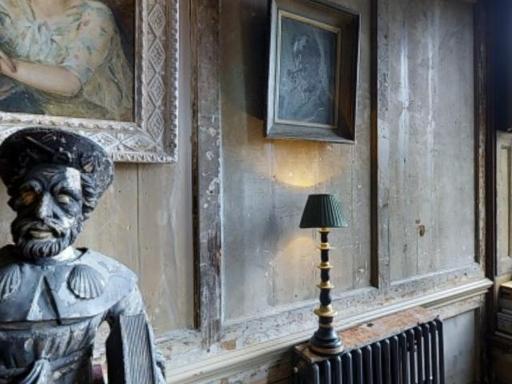} &
            \includegraphics[width=0.1\textwidth]{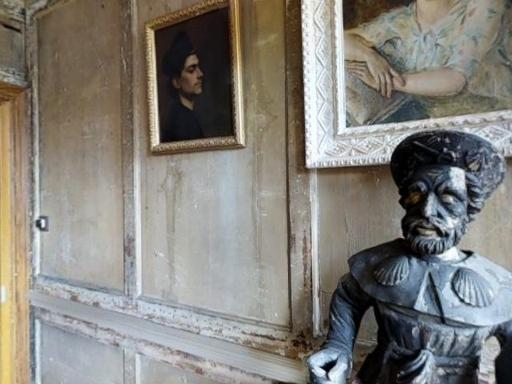} &
            \includegraphics[width=0.1\textwidth]{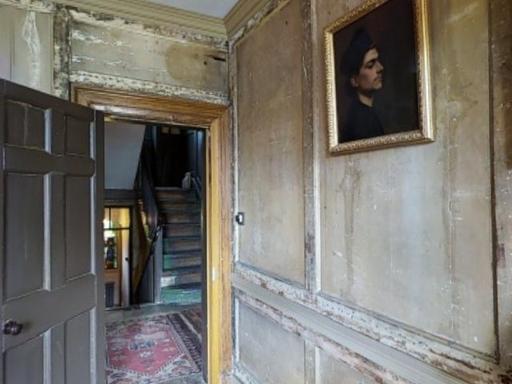} &
            \includegraphics[width=0.1\textwidth]{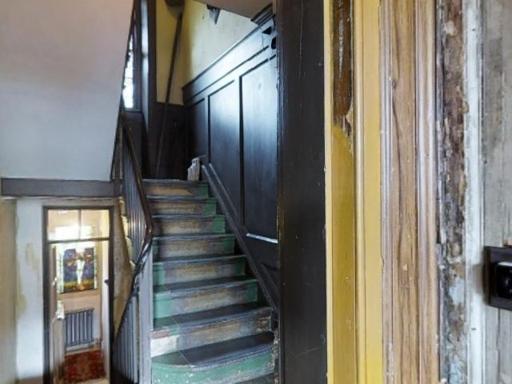} &
        \end{tabular}
    }
        \resizebox{.9\linewidth}{!}{
    \centering
        \begin{tabular}{cp{140mm}}           
            \addlinespace[-2pt]
           \multirow{2}{2cm}{\raggedleft \textbf{GT}:} & \textit{Okay. Left, right, center left, or center right?}
           \newline
           \textit{Take the right narrow doorway and look for more stairs that continue down. Take them all way to the bottom.} \\
           \addlinespace[4pt]
           \multirow{2}{2cm}{\raggedleft \textbf{UNMuTe}:} & \textit{Should I go to the left or right side of the room?}
           \newline
           \textit{Make a right and head into the hallway and then make a right into the stairs} \\
           \addlinespace[2pt]
        \end{tabular}
    }
    \caption{Sample paths taken from the CVDN ``val unseen'' split, together with the corresponding ground-truth interactions and generated ones. The number of depicted steps has been artificially reduced to $6$ to facilitate the graphical presentation. We only show the frontal image of the panoramic observation at each timestep.}
    \label{fig:qualitative}
\end{figure*}

\textbf{Experiment using Different Constant Thresholds.} 
We also performed experiments considering different constant threshold values in comparison to the model using the learnable threshold. Considering the action probability distribution of the navigator when the questions are and are not asked in Fig.~\ref{fig:plots} of the main paper, we set the threshold to $0.9$, $1.0$, and $1.1$ choosing values that separate the two distributions. However, looking at the results in Tab~\ref{tab:fixed}, \ours with a learnable threshold value performs better than all the baselines using fixed threshold values with a minimum improvement in terms of goal progress of $0.36$ meters.

\subsection{NDH Task}
The navigation experiments of \ours are complemented with experiments on the NDH task. NDH consists of navigation episodes using dialogue instances as textual input. To this end, the dialogue annotations and the trajectories of CVDN are split to form a total of 7K navigation episodes.
Before training the navigation model for the task, we generate question-and-answer pairs using our dialogue model for each trajectory in the training split of NDH. Consequently, we train DUET on the resulting double-sized dataset, augmented with synthetically generated dialogues. 

As it can be seen from Tab.~\ref{tab:ndh_val_unseen}, we achieve state-of-the-art results on both ``val unseen'' and ``test unseen'' splits of NDH. In particular, \ours trained on the trajectory performed by the human annotator (Player) achieves goal progress of $5.88$ and $5.75$ for the ``val unseen'' and ``test unseen'' respectively. \ours trained on the shortest path trajectory (Planner), instead, achieves a SPL and SR of $49$ and $60$ on ``val unseen'' and of $47$ and $56$ on ``test unseen''. The high difference in the SPL and SR of the agents trained on the planner and player trajectories is due to the fact that the agent uses the shortest path annotation in the case of the planner trajectory. Instead, the player trajectory often includes mistakes and reconsiderations, thus requiring the agent to backtrack to a previously visited node and lowering the values of SPL. In the table, the first section comprises studies that employ ground-truth dialogue annotations as instruction. These do not generate their own dialogues but simply use the dialogue provided in the NDH task for navigation. The second section, instead, reports methods that generate additional synthetic dialogues. Overall, \ours achieves top-1 performance on all metrics of the NDH task. 


\subsection{Dialogue Generation}
In this section, we discuss the capability of our dialogue model to generate proper question-and-answer pairs. To this aim, we compare the generated questions and answers with human annotations using NLP and reference-based description metrics like BLEU~\citep{papineni2002bleu}, ROUGE~\citep{lin2004rouge}, METEOR~\citep{banerjee2005meteor}, CIDEr~\citep{vedantam2015cider}, and SPICE~\citep{spice2016}. Results are reported in Tab.~\ref{table:description_metrics}. Here, the question is asked by the ``navigator'' (upper portion of the table) and the answer is given by the ``oracle'' (lower part of the table). For calculating different metric scores, we compare the predicted sentences with the ground-truth ones in terms of their n-grams (\ie~a sequence of $n$ consecutive words). BLEU, METEOR, and ROUGE are commonly used for the task of evaluating translation and summarization, while CIDEr and SPICE have been specifically designed for the task of image description and are also employed in VLN works in which synthetic instructions are generated \citep{stefanini2022show}. As can be seen, most of the metric values are above $0.20$ for generating an answer close to the ground-truth answer, which outlines the linguistic capabilities of our model. We further notice that the metric values for the ``navigator'' role are lower than those of the ``oracle'', \ie~the model is better at generating correct answers rather than asking proper questions. This is because there can be greater diversity in the generation of a question than that of the answer, which is instead more objective and should match the actions in the given trajectory.

\textbf{Future Images for Answer Generation.}
We then validate the contribution given by the incorporation of images extracted from the future trajectory (\ie~$(v_{t+1}, ..., v_{t+k})$ in Eq.~\ref{eq:gpt}) during the generation of answers in the dialogue model. This is done by comparing \ours with the answers of a dialogue model trained without using future images. 
The results are provided in the lower part of Tab. \ref{table:description_metrics}. 
Comparing the two oracles we can observe that, 
the oracle that does not employ future images undergoes a drastic reduction in performance on the ``val unseen'' split. In fact, the CIDEr score in ``val unseen'' decreases from $0.179$ to $0.111$. Overall, this underlines the effectiveness of employing future frames as a conditioning signal for the dialogue model. 

\textbf{Target object for Answer Generation.} 
We also validated the contribution given by the target object (\ie~$o_{tgt}$ in Eq.~\ref{eq:gpt}) during the generation of answers in the dialogue model. For this, we compared \ours with the answers of a dialogue model trained without using the target object. 
The results are provided in the lower part of Tab. \ref{table:description_metrics}. 
We can observe that, 
the oracle without the target object undergoes a reduction in performance on the ``val unseen'' split. The CIDEr score in ``val unseen'' decreases from $0.179$ to $0.145$. Overall, this shows that employing the target object as a conditioning signal for the dialogue model is beneficial for the generation of the answers. 

\subsection{Qualitative Generation Samples}
To showcase the quality of the proposed approach, we report three examples of generated dialogues in Fig.~\ref{fig:qualitative}. For all three examples, the question and answer generated by \ours appropriately describe the path that the agent should take. Noticeably, even if the ground-truth answer annotation of the first sample contains a mistake (the instruction is asking the agent to turn \text{left} rather than turning \textit{right}), \ours generates a correct answer, by asking the agent to turn right towards the living room. The second example consists of a yes-or-no interaction where the agent answers affirmatively to go down the long hall. In the third example, the agent asks a reasonable question on whether it should go right or left and the answer is clear and concise: go right, head into the hallway, and take a right to the stairs. As can be observed, these examples outline the effectiveness of the dialogue model and its ability to generate appropriate questions and answers for a given sequence of images.

\section{Conclusion}

This paper presents a novel computational model that engages in dialogue while navigating. The proposed architecture consists of a dialogue model and a navigator model: a fine-tuned GPT-2 decoder produces synthetic dialogues, and the navigation is predicted using a modified DUET model. 
The GPT-2 decoder is a multimodal text generator trained to generate questions using as input the target object and the current observation of the agent, while answers include future images along the trajectory to the goal. The modified DUET model is then trained to navigate using both ground truth annotation and generated dialogues.

Further, we learn an entropy-based ``whether-to-ask'' policy by minimizing a binary cross-entropy loss that predicts when it is beneficial to generate new dialogues. As a result, \ours learns to navigate more efficiently. 
We validated the effectiveness of our approach by performing extensive experiments triggering the dialogue model under different policies and settings. The final model achieves state-of-the-art performance on the most common Vision-and-Dialogue Navigation (VDN) datasets. 

In future work, we aim to assess the effectiveness of \ours by employing a human-in-the-loop methodology. This involves presenting future trajectory images to humans, who are asked to provide answers to the agent's questions. Additionally, exploring an object-based interaction, where humans inquire about the location of specific objects and the agent provides guidance on reaching them, could be another interesting extension of our work. However, this would necessitate a substantial adaptation of the proposed model and falls beyond the scope of the current study.

\bibliographystyle{abbrvnat}
\bibliography{biblio}

\end{document}